\documentclass[letterpaper, 10 pt, journal, twoside]{IEEEtran}
\ifCLASSINFOpdf
\else
\fi

\usepackage{amssymb}

\usepackage{amsmath}
\usepackage{algorithm}%
\usepackage{algorithmicx}%
\usepackage[noend]{algpseudocode}%
\usepackage{booktabs}
\usepackage{graphicx}
\usepackage{multirow}
\usepackage{lscape}
\usepackage{dblfloatfix}
\usepackage{titlesec}
\usepackage[hidelinks]{hyperref}
\usepackage{tikz}
\usepackage{mathtools, nccmath}
\usepackage{titlesec}

\begin{document}
\setlength{\parskip}{0pt}
%
\title{Bag of Views: An Appearance-based Approach to Next-Best-View Planning for 3D Reconstruction}
%
%
%

\author{Sara Hatami Gazani$^{1}$,
        Matthew Tucsok$^{2}$,
        Iraj Mantegh$^{3}$,
        and Homayoun Najjaran$^{1}$
\thanks{Manuscript received: July 12, 2023; Revised: October 11, 2023; Accepted: October 30, 2023.}
\thanks{This paper was recommended for publication by Editor Pauline Pounds upon evaluation of the Associate Editor and Reviewers’ comments. The authors would like to acknowledge the funding from the National Research Council (NRC) Canada under the grant agreement DHGA AI4L-129-2 (CDB \#6835).}%
\thanks{$^{1}$S. Hatami and H. Najjaran are with the Department of Mechanical Engineering, University of Victoria, Victoria, BC, CA, V8P 5C2 (e-mail: sarahatami@uvic.ca; najjaran@uvic.ca).}%
\thanks{$^{2}$M. Tucsok is with the Okanagan School of Engineering, University of British Columbia, Kelowna, BC, CA, V1V 1V7 (e-mail: tucsok@student.ubc.ca).}
\thanks{$^{3}$I. Mantegh is with the National Research Council (NRC) Canada, QC, CA (e-mail: iraj.mantegh@cnrc-nrc.gc.ca).}
\thanks{The authors have provided supplementary material (code and video) available at \href{https://github.com/ACIS2021/ViewPlanningToolbox}{https://github.com/ACIS2021/ViewPlanningToolbox} and \href{https://youtu.be/RhK2_HhaJoo}{https://youtu.be/RhK2\_HhaJoo}.}%
\thanks{Digital Object Identifier (DOI): see top of this page.}%
}

%
%

\markboth{IEEE ROBOTICS AND AUTOMATION LETTERS. PREPRINT VERSION. ACCEPTED OCTOBER, 2023}%
{HATAMI GAZANI \MakeLowercase{\textit{et al.}}: An Appearance-based Approach To Next-Best-View Planning for 3D Reconstruction}

%



\maketitle


\begin{abstract}
UAV-based intelligent data acquisition for 3D reconstruction and monitoring of infrastructure \textcolor{black}{has experienced} an increasing surge of interest due to recent advancements in image processing and deep learning-based techniques. View planning is an essential part of \textcolor{black}{this task} that dictates the information capture strategy and heavily impacts the quality of the 3D model generated from the captured data. Recent methods have used prior knowledge or partial reconstruction of the target to accomplish view planning for active reconstruction; the former approach poses a challenge for complex or newly identified targets while the latter is computationally expensive. In this work, we present Bag-of-Views (BoV), a fully appearance-based model used to assign utility to the captured views for both offline dataset refinement and online next-best-view (NBV) planning applications targeting the task of 3D reconstruction. With this contribution, we also developed the View Planning Toolbox (VPT), a lightweight package for training and testing machine learning-based view planning frameworks, custom view dataset generation of arbitrary 3D scenes, and 3D reconstruction. Through experiments which pair a BoV-based reinforcement learning model with VPT, we demonstrate the efficacy of our model in reducing the number of required views for high-quality reconstructions in dataset refinement and NBV planning.
\end{abstract}

\begin{IEEEkeywords}
Reactive and sensor-based planning, aerial systems: perception and autonomy, intelligent data acquisition.
\end{IEEEkeywords}

%
\IEEEpeerreviewmaketitle

\section{Introduction} 
 \begin{figure}[t]
\centerline{\includegraphics[width=\linewidth]{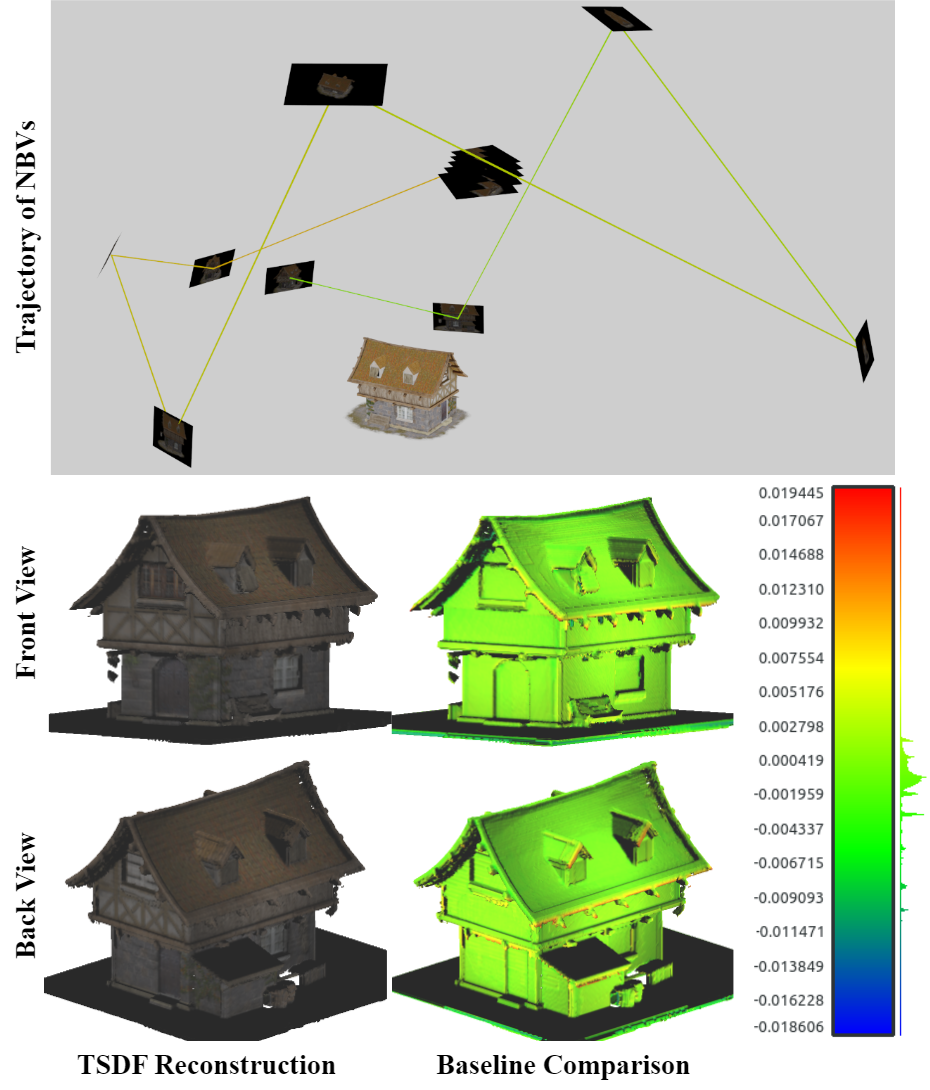}}
\caption{Results of the proposed Bag-of-Views model applied to next-best-view planning trained and evaluated using our View Planning Toolbox. The output reconstruction shows a $94.6\%$ surface coverage by only 13 captured views with $2.78$ cm Hausdorff distance and $0.80$ cm Chamfer discrepancy as compared to the results from a complete coverage baseline scan of the target with 288 views. }
\label{evalfig}
\vspace{-1.5em}
\end{figure}

\IEEEPARstart{A}{ctive} vision is characterized by the ability of a robot to make decisions about placing or reconfiguring its sensors to complement its perception of the environment~\cite{oldestsurvey}. This ability leads to meaningful actions of the robot based on interpretations of its surrounding environment and grants the robot a planning strategy, namely view planning, for actively replacing its sensors to uncover most amount of information about the target. In the context of active 3D reconstruction, view planning is used to optimize the robot's path until the task requirements are satisfied. For the application of 3D reconstruction of infrastructure using UAV-based imaging which is the concern of this work, the view planning problem dictates the data acquisition process and significantly impacts the reconstruction results. Previous work in this domain either relies on a given proxy of the target to build upon while planning the views~\cite{scott2009model, aryan2021planning} or generates a partial reconstruction using the knowledge of the agent about the target so the camera is navigated to complete the model~\cite{3dfrontier, rlpolicyrecon}. In this setting, the agent iteratively calculates the next waypoint to attend where it can capture the next-best-view (NBV) with the highest predicted information gain. However, when the purpose is to capture views from newly recognized targets or in the case of targeting complex structures, a geometric proxy of the target might not be accessible and online 3D reconstruction to achieve guidance can be computationally expensive and time-inefficient. On the other hand, under the assumption that adequate computation resources exist onboard the drone, algorithms that use an external model for guidance purposes mostly focus on the coverage completeness of the area where less attention is paid to the relative visual information contained in consecutively captured views. 

In this work, we propose a novel approach to a fully appearance-based view planning for online NBV planning and offline dataset refinement applications. The key uniqueness of our work lies on its model-free nature that makes it independent of the true state of the environment, namely the actual 3D model, both during training and at inference time. This allows the method to be applied to a wide range of settings and customized to fit different applications. In order to introduce the concept, we draw parallels between computational representation of views and how humans perceive objects by recognizing and interpreting distinct visual cues~\cite{torralba2006contextual}. We employ the Bag-of-Visual-Words~\cite{bovog} as a simplified vision model to represent an agent's knowledge of the target and introduce spatial information to the visual vocabularies to form a Bag-of-Views (BoV), a model to record and retrieve the visual features of the target from different viewpoints. \textcolor{black}{In the context of the application explored in this study where UAV flight duration is restricted due to power consumption concerns, the agent's task is to identify viewpoints that avoid having the camera repeatedly traverse back and forth around the target. Instead, it should chart a continuous and consistent path around the target.
In summary, our approach is distinguished by two distinctive attributes: i) it does not necessitate real-time reconstructions during training or inference, and ii) it ensures a smooth and uninterrupted planned trajectory.} First, through experimental results, we demonstrate how the selection of the views using our appearance-based heuristics affects the 3D reconstruction process and use the observations to refine already acquired datasets. Next, we bring this approach to an active level and utilize a Soft Actor-Critic (SAC) method~\cite{sacog} to train an agent that seeks to capture the target from views that cause a more drastic change in how it remembers the environment so far. The core idea of this method is to use the local visual features of the scene and their positioning to guide the agent to predict the NBVs that would result in revealing the highest number of unseen visual features as the agent remembers them. In addition, we present \textcolor{black}{the} View Planning Toolbox (VPT) as a comprehensive solution for simulating the view capturing process as part of developing view planning algorithms. The introduction of this toolbox fills \textcolor{black}{a} significant gap in the area of simulating machine learning-based view planning frameworks and provides developers with an open-source, user-friendly solution to streamline their training and evaluation process.
The main contributions of this work can be summarized as follows:
\begin{itemize}
    \item Bag-of-Views, a novel fully appearance-based view selecting model for offline dataset refinement. This model needs no pre-training and is modular and customizable for different applications. 
    \item A reinforcement learning approach to appearance-based NBV planning with no complete or partial reconstructions required \textcolor{black}{during training or at inference time.} 
    \item View Planning Toolbox (VPT), which provides an environment for training and testing of machine learning-based view planning and 3D reconstruction models.
\end{itemize}
\vspace{-0.8em}
\section{Related Work}\label{relatedwork}
\subsection{Solutions to the View Planning Problem}
Based on the amount of available information to the system about the environment and \textcolor{black}{visual complexity} of the target, view planning approaches are often categorized as model-based or model-free (a.k.a. non-model-based) approaches~\cite{maboudi2022review}. 
In model-based view planning problems, a viewing plan is obtained using a previously built or given model of the target~\cite{scott2009model, aryan2021planning}. 
In more general applications and in cases where the target is unknown or introduced to the system \textcolor{black}{at} runtime, the viewing strategy should be generated without prior information of the target~\cite{oldestsurvey}. In such cases, the goal is to manipulate the camera position and orientation in a manner that most unrevealed information about the target is exposed to the agent in each step. Most model-free methods follow the NBV approach. 
Typically, these systems build an interpretation of the environment using acquired information and base the planning of the next view(s) on it. That interpretation of the scene can be in different forms based on the specific task and application. Among methods that follow this approach are \textcolor{black}{frontier-based and} volumetric-based methods. 
In frontier-based view planning first introduced in~\cite{firstfrontier}, the core idea was to visit unexplored regions of an initial map, i.e. frontiers, to update the map based on the newly collected information in those regions. Methods belonging to this category usually represent the target zone of the environment using a 2D~\cite{2dfrontier} or 3D~\cite{3dfrontier} occupancy grid or directly use a point cloud to map the boundary between explored and unexplored regions~\cite{directfrontier}. As opposed to frontier-based methods that mostly focus on exploring an environment, volumetric-based approaches are usually concerned with modelling a single target and focus more on the completeness of the coverage. In this regard, to guarantee a successful registration, overlaps between the views are also considered while planning the views~\cite{vasquez}. In addition to 3D modelling, such algorithms are also used for scene inspection~\cite{voluminspection, heng}. 
A subcategory of the model-free view planning algorithms, namely appearance-based planning methods, carries out the decision making process based solely on the visual inputs such as RGB or \textcolor{black}{grayscale} images~\cite{appbasedrecons2014, rlpolicyrecon, scanning}. These methods either rely on an \textit{a priori} model of the target or a partial reconstruction of the scene, at least during the training stage of their models. For example,~\cite{appbasedrecons2014} proposed a method to compute a candidate sequence of viewpoints for a micro aerial vehicle to attend based on the information gain from different camera poses. More recently, with the advancements of deep reinforcement learning, appearance-based view planning has been visited more often. Accordingly,~\cite{rlpolicyrecon} utilized only the captured images to plan the views without tracking a partial reconstruction of the true 3D model. They used the surface coverage percentage to guide the agent. Similarly,~\cite{scanning} used the surface coverage as well as reconstruction error as part of the reward for guiding their reinforcement learning agent towards complete reconstruction. Unlike these methods, our proposed model omits the need for the true state of the environment or any partial reconstructions for guiding the agent towards capturing high-utility views for the task of reconstruction. We treat 3D reconstruction as a downstream task as opposed to a parallel task in the view planning framework and pay more attention to the conditions that should be met by the views in order to have a satisfactory reconstruction.
\vspace{-1.0em}
\subsection{Simulation Environments for The View Planning Problem}
View planning research has explored various simulation environments, each exhibiting unique advantages and constraints. For instance, the PredRecon framework \cite{predrecon} utilized AirSim \cite{airsim} for its simulation environment, taking advantage of its high-fidelity simulation capabilities. However, their work also relied on Unreal Engine 4 (UE4) \cite{unrealengine} for data generation of other 3D models and Blender \cite{blender} for partial pointcloud reconstructions. This use of multiple software environments can be resource-intensive and complex to set up. In a similar manner, \cite{nbv_large3d_multi_uav} used Gazebo \cite{gazebo}. While Gazebo is a feature-rich simulator providing a versatile environment for robotic simulations, using it for view planning in photo-realistic environments introduces computational overhead and performance degradation.

Previous work addressed the need for flexibility and customization by employing environments inspired by OpenAI Gym \cite{openai_gym} to train their reinforcement learning agent for NBV planning~\cite{scanning}. Their approach offered greater flexibility in defining the simulation environment and incorporating different perceptual elements. Additionally, the \textit{gym-collision-avoidance} package \cite{gym_collision_package} used in \cite{Where_to_look_next} provided a simulation environment for informative trajectory planning. With a greater focus on collision avoidance, this package allowed researchers to simulate and evaluate view planning algorithms within a controlled environment.


It is worth noting that simulation environments often require dealing with trade-offs between fidelity, computational resources, and the level of control provided over environmental factors. Striking the right balance between these aspects is dependent on the specific task being studied. For view planning, factors affecting high-level control of the simulation are of most importance. These factors include camera controls (pose, focal length, resolution), scene manipulation (lighting, object placement and scaling), and scripting tools for automating the aforementioned functionalities. To address these trade-offs, we have developed the View Planning Toolbox (VPT), a comprehensive solution outlined in Sec.~\ref{env}.
\vspace{-0.8em}
\section{Bag of Views: Appearance-based Approach for Selecting Best Views} \label{bovsec}
Following the work in~\cite{ours}, we introduce a computational representation of the views in terms of the visual features of the scene captured from the respective viewpoints. As discussed in~\cite{ours}, two key conditions must be satisfied for a successful multi-view 3D reconstruction:
\begin{itemize}
  \item The views in an input set must present features of the target that are distinct from the ones presented by other views in the same set,
  \item Each view by itself must be rich in the number of visual features it is revealing of the target. 
\end{itemize}

We denote the $i^{th}$ view in the set as $\chi_i$. Each view $\chi_i$ is encoded to and is represented by its extracted features using a feature extracting algorithm such as Scale-Invariant Feature Transform (SIFT)~\cite{Lowe04distinctiveimage}. Thus, $\chi_i$ will be a 2D matrix with dimensions $m\times n$ where $m$ is the number of detected keypoints and $n$ is the number of values in the feature descriptors. In the case of SIFT, each row of this view matrix represents a 128-dimensional feature descriptor where each element represents a certain attribute of the local feature detected in the image patch. We can denote each view by iterating over its resulting feature descriptors as $\chi_i(j, :) = \{f(j,k): k\in\{1,2,...,n\}\}$ where $f(j,k)$ is the $k^{th}$ value in the $j^{th}$ descriptor of the image. These conditions for a set of views result in a greater distance between corresponding descriptors from two view representations denoted as  $dist(\chi_i(j,:), \chi_{i+1}(j',:))$ for consecutively selected random views $\chi_i$ and $\chi_{i+1}$. A greater distance ensures a better reconstruction quality for a limited-length trajectory. In our case, the cosine distance metric is used to measure the dissimilarity between the descriptors and the visual words since its consistent range of outputs allows for easy interpretation and score comparison. Thus,
\begin{equation}
    dist(\chi_i(j,:), \chi_{i+1}(j':)) = 1 - 2\times \cos{(\chi_i(j,:), \chi_{i+1}(j',:))} 
\end{equation}
where $j \in \{1,2,...,m\}$ and $j'\in \{1,2,...,m'\}$ with $m$ and $m'$ being the number of representative descriptors of $\chi_i$ and $\chi_{i+1}$:

\begin{equation}
   \cos{(\chi_i(j,:), \chi_{i+1}(j':)(j))} = \frac{\langle {\chi_i(j,:)} , \; {\chi_{i+1}(j',:)}\rangle}{|\chi_i(j,:)|.|\chi_{i+1}(j',:)|}
\end{equation}

Since the extracted features belonging to a target are prone to self-similarity, the feature descriptors of different regions of the target can be clustered based on their similarity and, instead of pair-wise comparisons of all feature descriptors belonging to all views in the set, they can be compared to the cluster cores. In addition, the necessity of there being a model-free view planning \textcolor{black}{model} with no pre-training requires learning these cluster cores iteratively from the incoming information. This clustering of the visual features is inspired by~\cite{recognition} where cluster centers of the quantified feature descriptors were used to form a Bag of Visual Words (BoVW).


In the context of view planning for a single target with one or more symmetry axes, learning a global visual vocabulary for the entire model can be prone to overconfidence in recognizing certain visual words. Therefore, we propose a method to track visual features of the target captured from different viewpoints via distinguished visual vocabularies for different regions included in what we call a Bag-of-Views (BoV). As well as mitigating the symmetry challenge, the computation cost will be significantly less with the viewpoint-based vocabularies; a smaller number of visual words is required to describe a portion of the target rather than all parts of it and the new view will only update its corresponding vocabulary among the BoV.  Below are the steps involved:


\begin{algorithm}[b]
\caption{Bag-of-Views Model for Offline View Selection}\label{alg: alg1}
\begin{algorithmic}[1]
\Require Number of view ranges $N$, Number of words $W$, Database $\mathcal{D}$
\State \textbf{Initialize} BoV\{${\nu_{1:N}}$\}, Feature extractor $\mathcal{F}$
\While{There is data to process in $\mathcal{D}$}
    \State $T, X \gets$ Load data from $\mathcal{D}$
    \State $id_{\text{view}} \gets T.\text{azimuth} // \frac{2\pi}{N}$
    \State $\chi \gets \mathcal{F}(X_i)$
    \State $dist \gets 0$
    \For{$j = 1$ to $W$}
        \State $C \gets \underset{k}{\arg\max}\left(\cos\left(\chi(j), \nu_{id}(k)\right)\right)$
        \State $dist \gets dist + (1-2\times\cos\left(\chi(j), \nu_{id}(C)\right))$
        \If{$dist > 0$}
            \State Update descriptors with $\chi$ for $\nu_{id_{\text{view}}}$
            \State $\nu_{id_{\text{view}}}.codebook \gets$ Perform K-means on $\nu_{id_{\text{view}}}$
        \Else
            \State Remove $X$ from $\mathcal{D}$
        \EndIf
    \EndFor
\EndWhile
\end{algorithmic}
\end{algorithm}

\textbf{1) Feature Extraction:} Using a feature extraction algorithm, in our case SIFT, local features of the captured image at position $T$ are extracted in the form of 128-dimensional vectors. $T$ is the position of the camera in the spherical coordinate system with its origin located at the center of the scene.

\textbf{2) Utility Assignment:} Depending on the application at hand, assigning utilities to the views can be divided into two different cases:

\textbf{2.I)} In the case of dataset refinement where the views have already been captured, a decision should be made about including each view in the input set of the reconstruction algorithm based on its utility. The question simply is "does this new view help the reconstruction process?". To answer that, we look at the extent to which this view satisfies the two conditions \textcolor{black}{previously mentioned}. Each of the feature descriptors from the previous step are compared with their closest visual word through vector quantization~\cite{beyond}. Then, a distance metric is used to measure the dissimilarity between feature descriptors and the supposed visual word that would represent them in the corresponding vocabulary in the BoV, denoted as $\nu_{{id}_T}$. Here, ${id}_T$ is the identification number for the vocabulary associated with the region that $T$ belongs to. This process is repeated for all of the descriptors and the sum of the dissimilarity scores is used to decide whether to ignore or utilize the view in the reconstruction process. If the final score is a non-zero positive value, it is included in the set and proceeds to the third step, otherwise it is ignored.

\textbf{2.II)} In the second case, we use the BoV model to train a reinforcement learning agent to propose NBVs based on the appearance of consecutively captured views. 
Further elaboration on the learning process will be provided in Sec.~\ref{rl}. In this section, our focus lies primarily on the utilization of this model to shape the reward function within the specified context. 
While training the reinforcement learning agent, we use the change in the corresponding vocabulary of the BoV after capturing a new view at location $T$ to shape the reward function. \textcolor{black}{A greater} difference between the new and previous BoV implies that the new view contains more unseen features and results in higher rewards.

\textbf{3) View Representation:} This step is a continuation of case  $2.I$. Depending on the position and orientation of the captured viewpoint, the extracted features update the specific vocabulary assigned to the range of views that the new view belongs to. This updating includes clustering of the descriptors belonging to that region using a clustering algorithm such as K-means~\cite{kmeans}. Thus, every group of the cluster centers in the BoV, namely every regional vocabulary, describes the appearance of the target from viewpoints that are close in position and orientation.
Algorithm~\ref{alg: alg1} showcases the pseudo-code for the creating a BoV model.
\vspace{-0.5em}

\section{A Reinforcement Learning Approach to Appearance-based NBV Planning} \label{rl}
The problem of NBV planning is a sequential decision making process that can be defined as a Partially Observable Markov Decision Process (POMDP) and be solved through reinforcement learning algorithms. Our goal is to achieve this without any need for \textit{a priori} knowledge of the target and without any full or partial reconstruction of the target during training or inference time. We begin by formulating different components of this process. The goal of the agent in our system is to iteratively propose next views for a limited number of steps to reach regions with a high number of features unfamiliar to the BoV. Seeking such views leads to drastic changes in the vocabularies of the BoV through each relocation of the camera. 

The state space should provide the agent with enough information about the environment to enable meaningful actions towards the goal. We use the concatenation of down-sampled \textcolor{black}{grayscale} images captured through the last $\tau$ consecutive frames as well as the concatenation of the normalized camera locations associated with each view. Thus, the state $s_t$ at time $t$ is defined as $\{T_{t-\tau:t}, obs_{t-\tau:t}\}$. The camera location $T_t$ is presented using the spherical coordinate system in the form of $\{R, \phi, \theta\}$ with three values for radial distance from the center, the azimuth angle, and the elevation angle. Also, given a deterministic state transition, the action $a_t$ determines the next camera location $T_{t+1}$. \textcolor{black}{Depending on the policy network's architecture, $T_{t+1}$ may need to be adjusted to meet safety requirements. In our case, it is rescaled to conform to the specified ranges for its three components.}
Based on the introduction in case $I$ of Sec.~\ref{bovsec}, the reward received for this action represents the change in the part of the BoV that has been influenced by the new action, namely $\nu_{T_{t+1}}$ which is the vocabulary associated with the region that the new location belongs to. This change is measured \textcolor{black}{by} comparing the same regional vocabulary before and after taking the action; the closest visual words in the two vocabularies are identified and their distance is measured through vector quantization with the cosine distance metric. 
The sum of these distances is used to represent the change in the vocabulary after taking the action. We also add a negative constant reward at each time step. Thus, we define the reward to be $r_{t+1} = dist(\nu_{T_{t+1}}, \nu_{T_t}) - 1$, with the distance between vocabularies being defined as:
\begin{equation}
\text{dist}(\nu_{T_{t+1}}, \nu_{T_t}) =
\sum_i \left( 1 - 2 \times \cos \left( \nu_{T_{t+1}}(i), \nu_{T_{t}}(k) \right) \right)
\end{equation}
where $k=\underset{j}{\arg\max} \cos(\nu_{T_{t+1}}(i), \nu_{T_{t}}(j))$ and $i$ iterates \textcolor{black}{over each of the visual words} in $\nu_T$. The number of vocabularies in the BoV and the size of each vocabulary are dependant on the resources available during training and runtime.
\vspace{-0.5em}
\section{Simulation Environment} \label{env}
To address the need for a lightweight, flexible, and easy-to-integrate simulation environment, we introduce the View Planning Toolbox (VPT). VPT contains tools for simulating the components required to train, visualize, and test view planning algorithms entirely within Python \cite{python} and Blender \cite{blender}. \textcolor{black}{At the core of VPT is the \textit{UAV Camera}, which simplifies the visual data acquisition process carried out by an Uncrewed Aerial Vehicle (UAV) down to a camera floating in 3D space, where the specifications and pose of the camera can be determined either manually, programmatically, or through the use of a view planning algorithm to generate arbitrary 3D trajectories.}
\vspace{-1em}
\subsection{UAV Camera}
 \begin{figure}[t]
\centerline{\includegraphics[width=\linewidth]{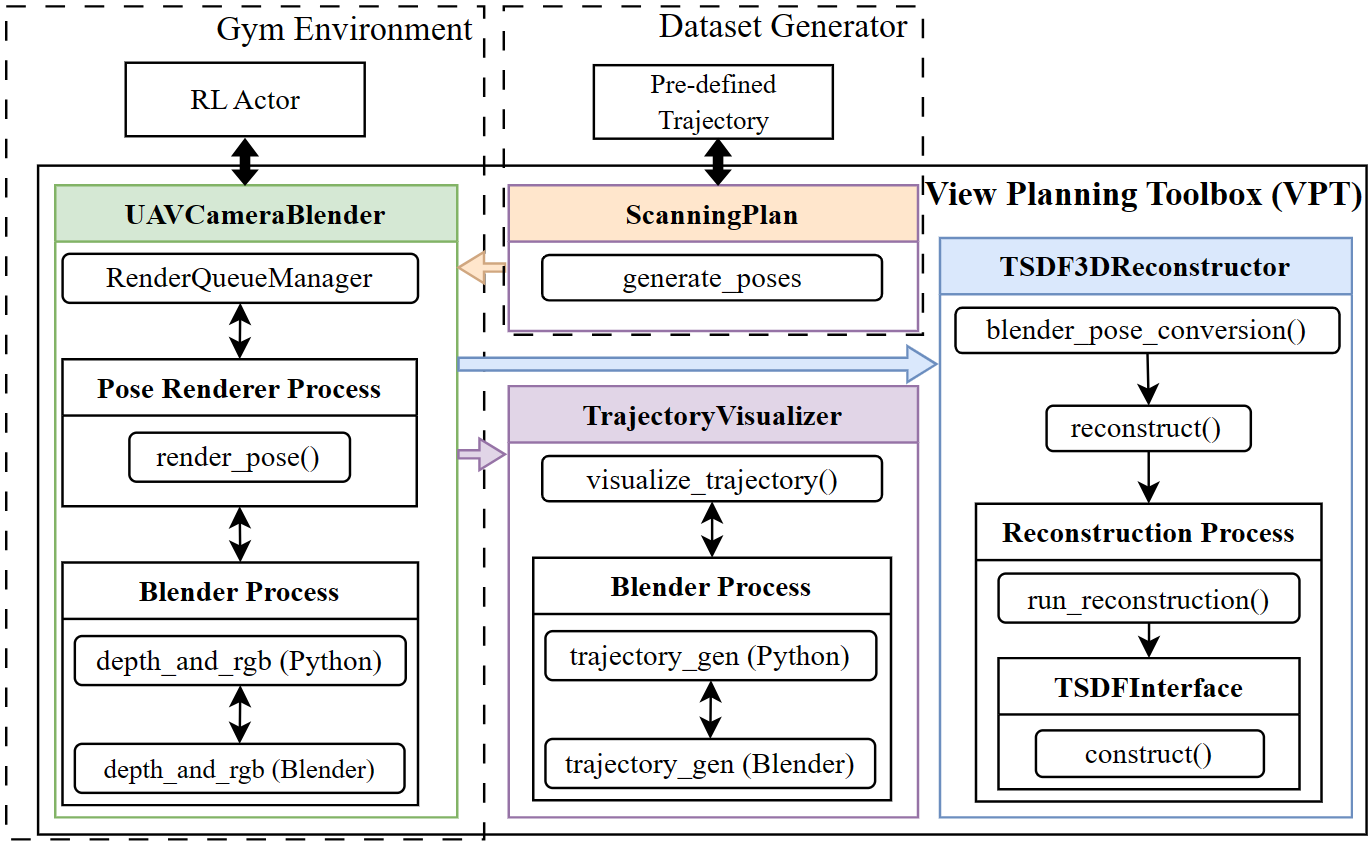}}
\caption{ Simplified visualization of the interactions between different components of the View Planning Toolbox (VPT) for reinforcement learning and dataset generation.}
\label{vpt_framework}
\vspace{-1em}
\end{figure}
 The \textit{UAV Camera} simulates the base functionality of a drone traversing a 3D scene to gather visual information including both RGB and depth images from the environment. It is a simple pinhole model where the developer specifies the resolution of the captured image in pixels, the focal length in millimeters, and the depth range in meters. The \textit{UAV Camera} implementation encapsulates BlendTorch \cite{blendtorch}, a framework designed to integrate PyTorch \cite{Pytorch} deep learning models with Blender \cite{blender}. The BlendTorch framework provides a convenient, multiprocessing-safe means of communication between a Blender environments and native Python applications. In addition, the \textit{UAV Camera} can be instantiated in an OpenAI gym \cite{openai_gym} environment for use with reinforcement learning models.
 \vspace{-1em}
\subsection{Data Generation}\label{gensec}
\begin{figure}[b]
\centerline{\includegraphics[width=\linewidth]{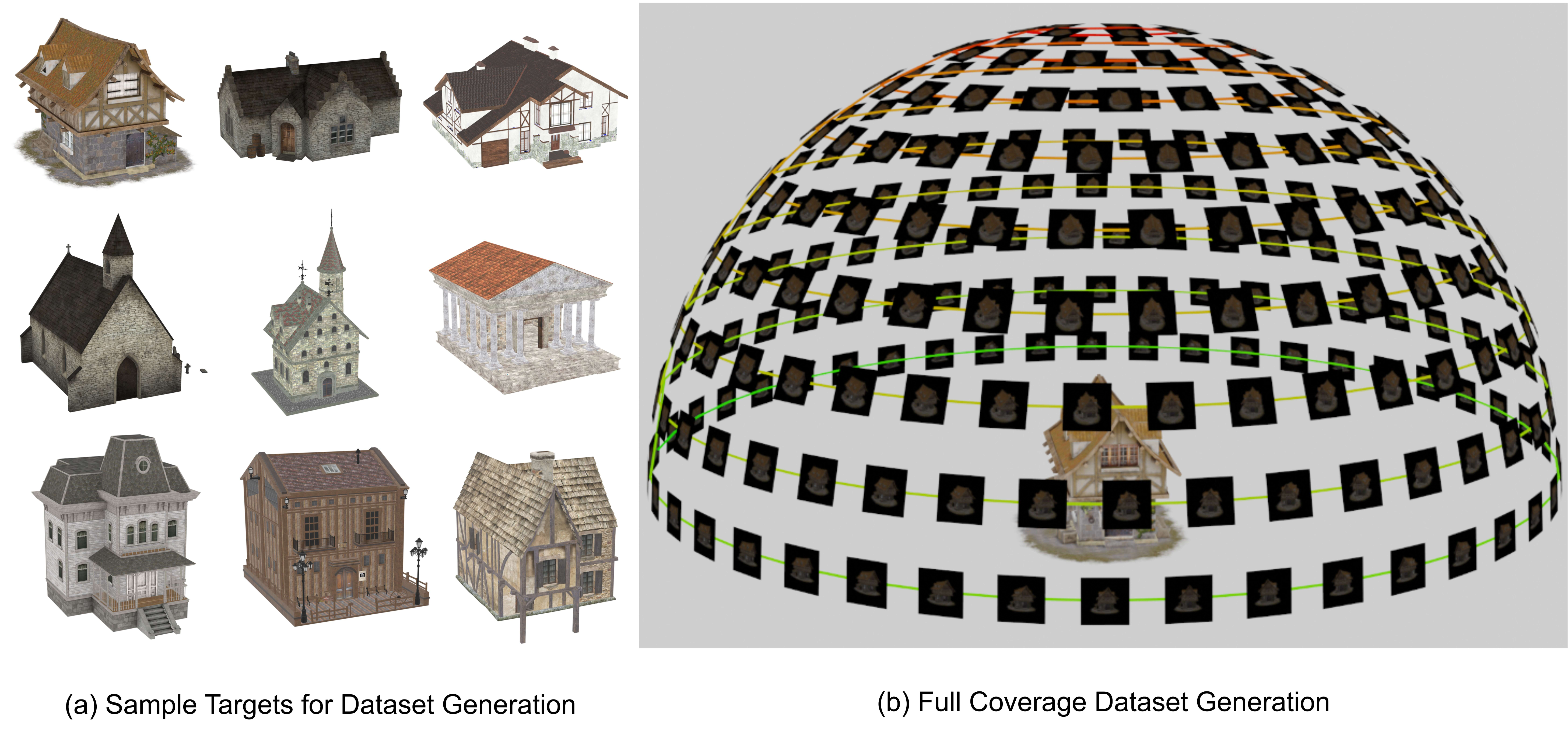}}
\caption{ VPT performs full coverage hemisphere scans in which the end and side overlap are specified on many targets.}
\label{data_gen}
\end{figure}

VPT can serve a dual purpose as shown in Fig. \ref{vpt_framework}:
\begin{itemize}
    \item Generate state-action pair experiences for training reinforcement learning models. This requires accessing the \textit{UAV Camera} directly which can be instantiated in a Gym environment.
    \item Generate offline datasets for training supervised learning models. This functionality is contained within \textit{Scanning Plans} and adheres to traditional photogrammetric principles~\cite{photogrammetry_basics}, ensuring proper determination of end and side overlap for aerial scanning.
\end{itemize}
As a way of evaluating our appearance-based view planning model in Sec.~\ref{bovsec}, we compare the resultant reconstructions from the captured views with those generated by views that fully cover the targets. \textcolor{black}{These views are generated using the  \textit{hemispherical scan} tool in VPT, which maintains a constant end and side overlap between images at a distance $R$ from the center of the scene. The spacing of views around the hemisphere are calculated by making a series of flat plane and small angle approximations due to the high overlap between views.} Each scene contains a photo-realistic structure which has been centered about the z-axis and lies on the XY-plane. The structure is also normalized by its largest dimension. This ensures structures of all sizes can be fully visible from all viewpoints. A resulting trajectory of views generated with this method can be seen in Fig. \ref{data_gen}. 
\vspace{-1em}
\subsection{TSDF-Fusion 3D Reconstruction}\label{tsdf}
To evaluate the performance of our view planning algorithms, VPT integrates a Truncated Signed-Distance Function (TSDF) Fusion 3D reconstruction module from \cite{3dmatch}. A wrapper has been implemented to isolate the CUDA environment of the TSDF-Fusion module to prevent conflicts with PyTorch. The TSDF-Fusion module creates either a 3D point cloud or a 3D mesh using the marching cubes method \cite{marching_cubes}. TSDF-Fusion reconstruction requires RGB views, the corresponding depth maps, and the computed $4\times4$ rigid transformation matrices based on the poses for the corresponding views. This module acts as a lightweight photogrammetric stand-in to quickly evaluate the reconstruction quality from hundreds of views.  
\vspace{-1.4em}
\section{Implementation Details and Experiments}\label{impexp}
\subsection{Network Architecture and Training}\label{training}
\begin{figure}[t]
\centerline{\includegraphics[width=\columnwidth]{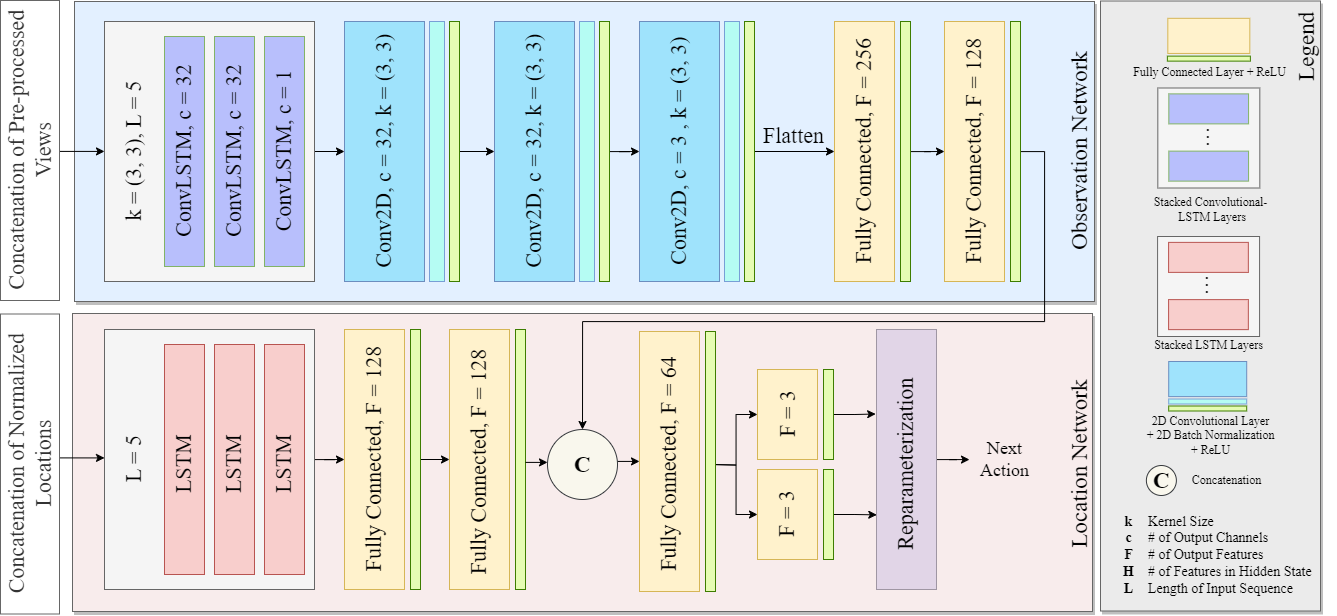}} 
\caption{Proposed architecture for the policy network. The policy network processes the two components of the state, namely the observations and locations throughout the last 5 frames, through two separate sub-networks and fuses the resultant feature vectors to produce the action distribution.}
\label{architecturefig}
\vspace{-1.2em}
\end{figure}

\begin{figure}[t]
\centerline{\includegraphics[width=\columnwidth]{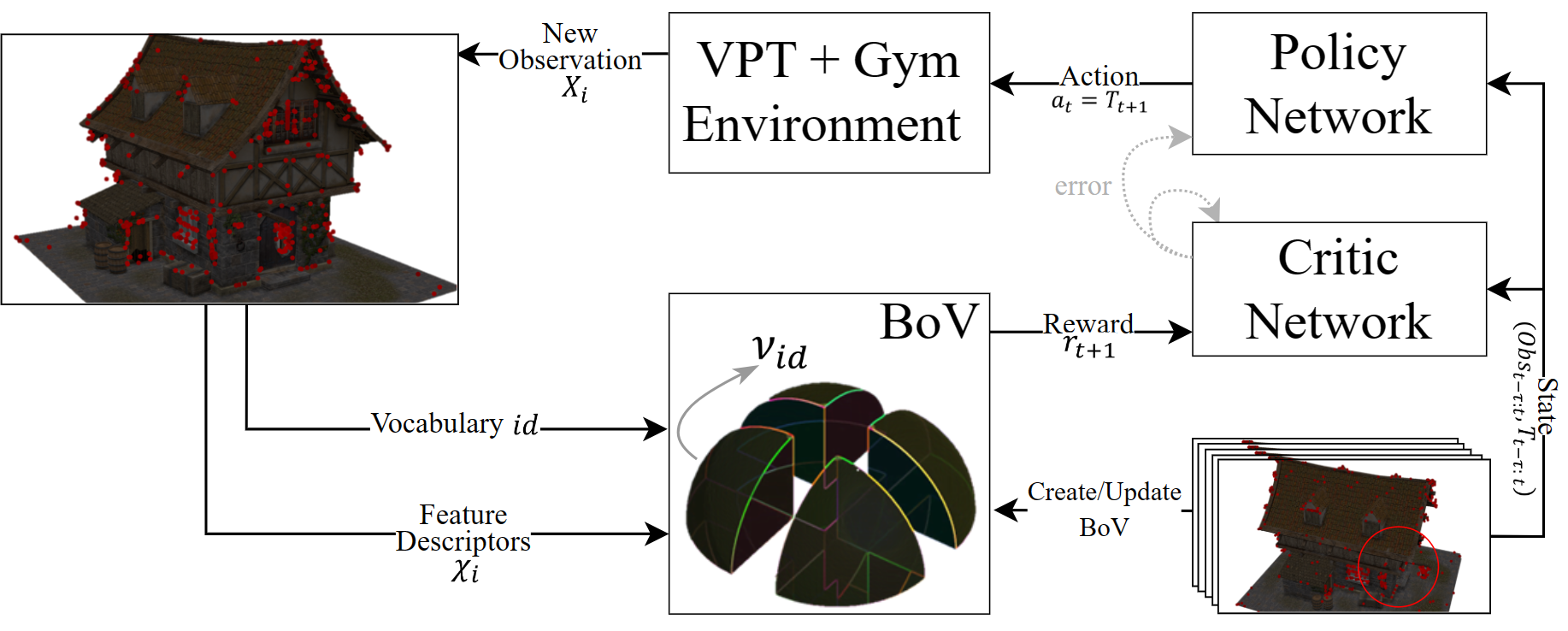}} 
\caption{Block diagram of the proposed reinforcement learning approach to NBV planning based on Bag-of-Views model.}
\label{nbvframework}
\vspace{-1.4em}
\end{figure}

The policy network is composed of two different sub-networks for processing the two components of the state; the observation network and the location network. Details of these two sub-networks are shown in Fig.~\ref{architecturefig}. 
We used the Soft Actor-Critic (SAC) algorithm by~\cite{sacog} to train the policy network. SAC is powered by the advantages of actor-critic methods and maximum entropy reinforcement learning. It is a model-free, off-policy algorithm that optimizes both the policy and the value function simultaneously through a combination of policy gradient and Q-value updates. It maximizes the expected return while incorporating a soft entropy regularization term~\cite{haarnoja2018soft}.
The critic network, which is responsible for mapping state-action pairs to their quality values, encodes the concatenation of the observation component of the state through a sequence of 3D convolutional layers followed by 3D batch normalization layers and ReLU activation functions. Then, the output is flattened and goes through two fully connected layers with an output size of 128. A sub-network also processes the concatenation of the location component of the state and the action through fully connected layers of output size 64 and 128, respectively. The resulting feature vector is then concatenated with the encoded observations and goes through two fully connected layers of output size 128 and 3 as the output quality value vector. \textcolor{black}{Training is done using the hyperparameters listed in Table~\ref{hyperparam} and a simplified version of the proposed reinforcement learning cycle is depicted in Fig.~\ref{nbvframework}. }
\vspace{-1.8em}
\begin{table}[hp]
\centering
\caption{\textcolor{black}{Table 1: Hyperparameters used for training the RL agent}}
\label{hyperparam}
\begin{tabular}{cc}
\hline
\textcolor{black}{Parameter} & \textcolor{black}{Value} \\ \hline
\textcolor{black}{Batch Size} & \textcolor{black}{8} \\
\textcolor{black}{Learning Rate} & \textcolor{black}{$5\times10^{-4}$} \\
\textcolor{black}{Temperature Parameter ($\alpha$)} & \textcolor{black}{0.2} \\
\textcolor{black}{Target Network Update Interval (steps)} & \textcolor{black}{20} \\
\textcolor{black}{Experience Replay Buffer Size} & \textcolor{black}{$10^6$} \\
\textcolor{black}{Soft Update Factor ($\tau$)} & \textcolor{black}{0.01} \\
\textcolor{black}{Discount Factor ($\gamma$)} & \textcolor{black}{0.99} \\ \hline
\end{tabular}
\end{table}
\vspace{-2em}


\begin{table*}[b]
\centering
\caption{Studying the influence of the size of Bag-of-Views on the dataset size and the reconstruction results from the TSDF fusion of RGB and depth images}
\label{bovtable}
\resizebox{\textwidth}{!}{%
\begin{tabular}{llllllllllllll}
\hline
\multicolumn{2}{l}{Size of BoV} & 5 & 5 & 5 & 7 & 7 & 7 & 9 & 9 & 9 & 11 & 11 & 11 \\ \hline
\multicolumn{2}{l}{Number of Words per View Range} & 10 & 20 & 30 & 10 & 20 & 30 & 10 & 20 & 30 & 10 & 20 & 30 \\ \hline
\multicolumn{1}{l|}{\multirow{3}{*}{Medieval Tavern}} & Number of Selected Views & 143 & 67 & 37 & 142 & 63 & 34 & 140 & 62 & 35 & 146 & 63 & 42 \\
\multicolumn{1}{l|}{} & Hausdorff Distance (cm) & 2.1509 & 1.8134 & 1.8901 & 1.8045 & 2.3907 & 2.2756 & 1.8803 & 2.3693 & 1.5216 & 1.4825 & 1.7841 & 1.4355  \\
\multicolumn{1}{l|}{} & Chamfer Discrepancy (cm)  & 0.1183 & 0.1935 & 0.4542 & 0.1206 & 0.4345 & 0.3742 & 0.1254 & 0.1906 & 0.2998 & 0.1109  & 0.1953 & 0.2979  \\ \hline
\multicolumn{1}{l|}{\multirow{3}{*}{Forest Ruin House}} & Number of Selected Views & 102 & 38 & 23 & 92 & 33 & 24 & 98 & 42 & 32 & 99 & 47 & 36 \\
\multicolumn{1}{l|}{} & Hausdorff Distance (cm) & 2.5810 & 2.2692 & 2.5231 & 2.0577 & 2.1886 & 3.1446 & 2.1244 & 2.2862 & 2.3698 & 2.04848 & 2.6412 & 1.7054  \\
\multicolumn{1}{l|}{} & Chamfer Discrepancy (cm) & 0.2374 & 0.4554 & 0.6548 & 0.2416 & 0.5375 & 0.5775 & 0.1869 & 0.6057 & 0.5842 & 0.3115  & 0.5359 & 0.5535  \\ \hline
\end{tabular}%
}
\end{table*}

\subsection{Evaluation of The View Planning Results} \label{eval}
To evaluate the utility of the suggested views using the BoV model, we used the reconstruction method explained in Sec.~\ref{tsdf} and compared the resulting 3D point clouds and meshes from the captured views with those from a complete coverage scan of the target discussed in Sec.~\ref{gensec}. We analyzed the reconstruction results of the offline dataset refinement and online NBV planning using the Chamfer discrepancy, Hausdorff distance, and the mesh-to-mesh comparison carried out in CloudCompare~\cite{cloudcompare3}. Chamfer discrepancy gives an overall measure of similarity or dissimilarity between two point sets, while Hausdorff distance focuses on the largest observed distance, highlighting extreme differences. \textcolor{black}{When comparing scanning results using these metrics, a higher Chamfer discrepancy indicates poorer coverage of the target and lower overall accuracy while a higher Hausdorff distance suggests that there are missed areas in the scanning process, leading to a reconstruction with holes.}
\vspace{-1.2em}

\subsection{Bag-of-Views for Dataset Refinement}
\begin{figure}[t]
\centerline{\includegraphics[width=\columnwidth]{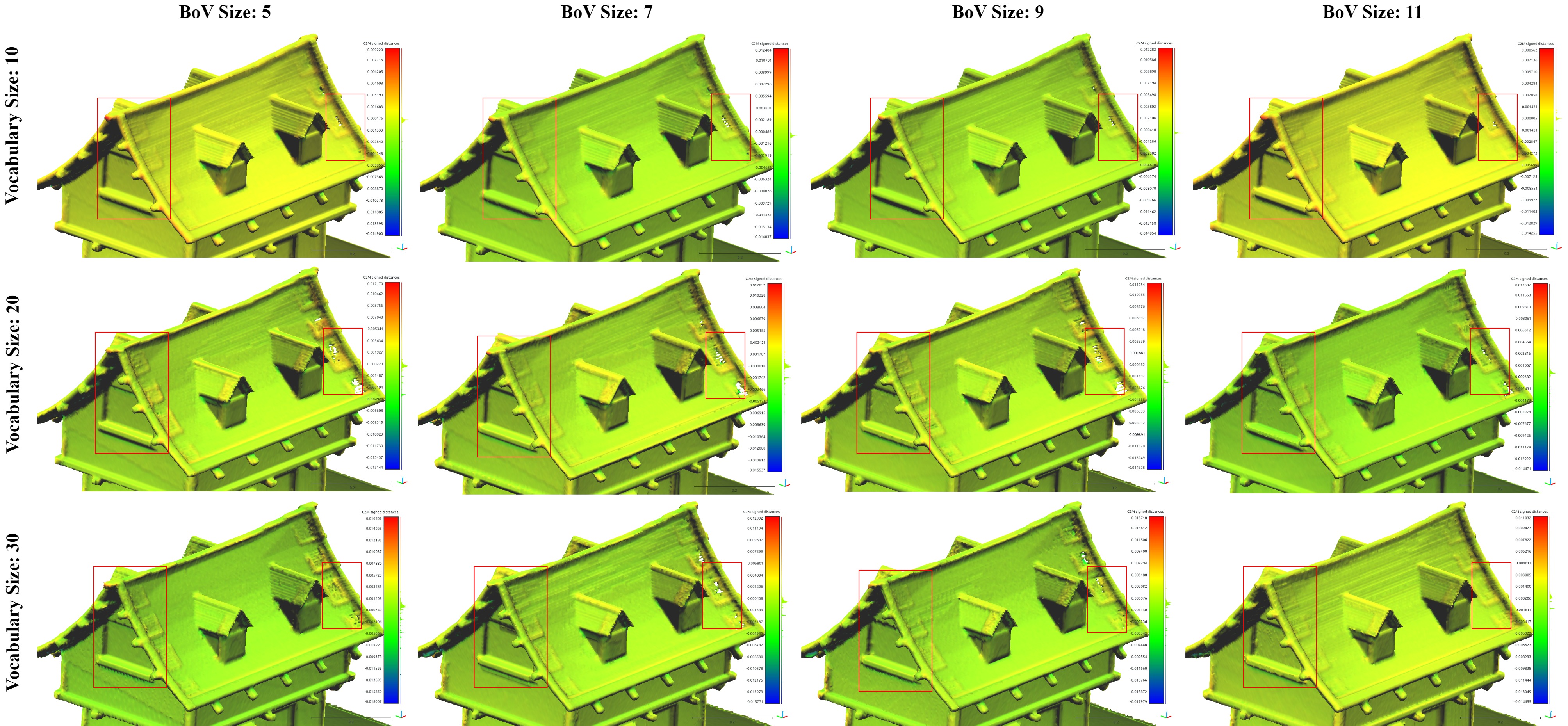}} 
\centerline{\includegraphics[width=\columnwidth]{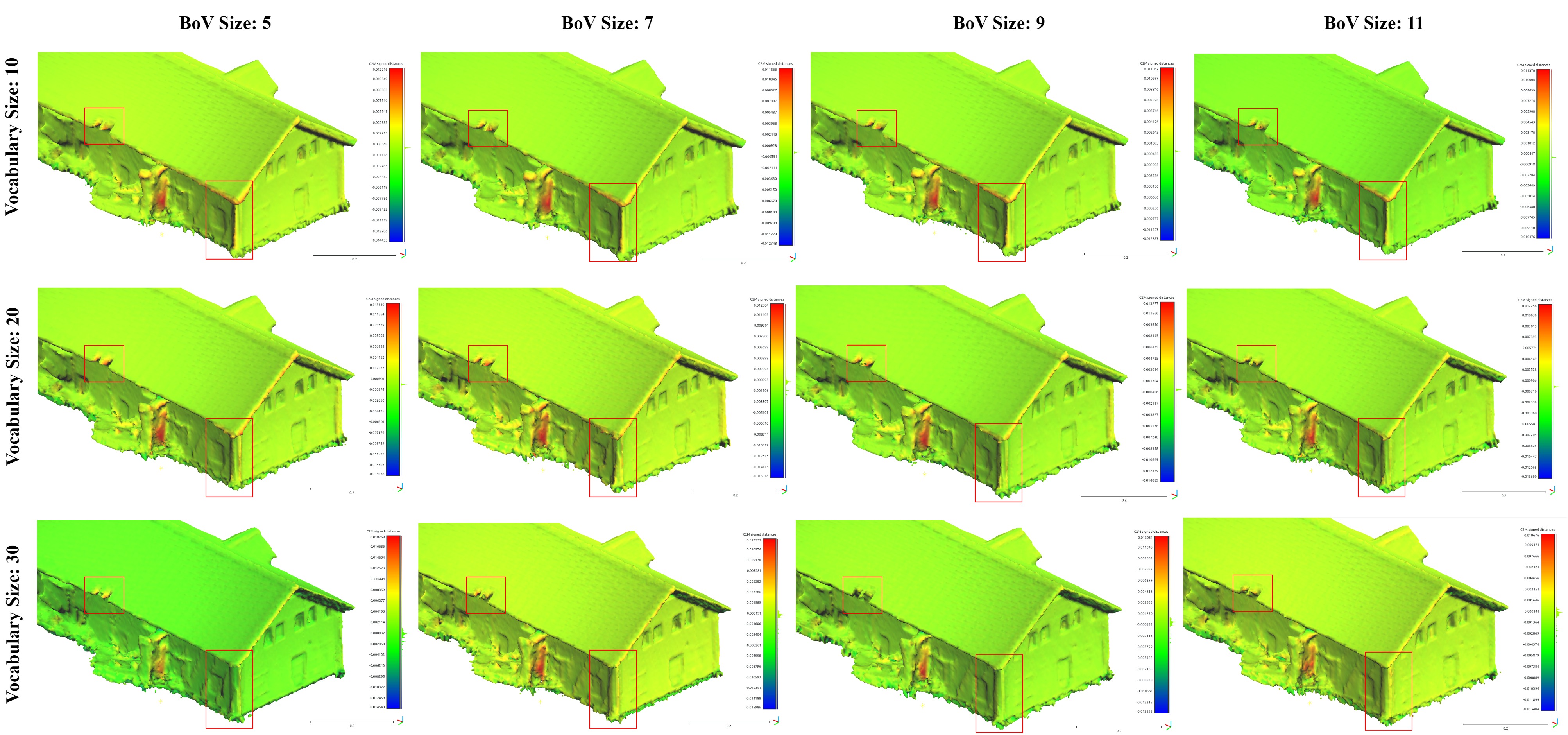}}
\caption{Studying the effect of the size of BoV and the vocabularies within it. It can been seen that increasing the size of BoV lowers the error sparsity while increasing the vocabulary size reduces the error values.}
\label{bovperf}
\vspace{-1.0em}
\end{figure}

To study the effect of the size of BoV, we \textcolor{black}{tested} the model performance in filtering two of the generated datasets consisting of 288 views uniformly covering two of the buildings in our dataset as explained in Sec.~\ref{gensec}. We used the method explained in Sec.~\ref{bovsec} and Alg.~\ref{alg: alg1} to reduce the dataset size and evaluate reconstruction performance resulting from the remaining views.
Two effective variables in this study \textcolor{black}{were} the number of view ranges defining the size of BoV and the number of visual words in each vocabulary of the BoV. The reconstruction results were compared based on their Hausdorff distance and Chamfer discrepancy with the 3D point cloud reconstructed using the original dataset of 288 views. The results are listed in Table~\ref{bovtable}. In addition to this quantitative comparison, we visualized the distance between the mesh reconstructions to those produced using the original dataset in Fig.~\ref{bovperf}.


This analysis \textcolor{black}{explored} the question of achieving optimal performance by balancing model efficiency (number of selected views) and reconstruction quality. Shown in Fig.~\ref{bovperf}, a larger BoV, achieved by increasing the number of vocabularies, \textcolor{black}{resulted} in a reconstruction that exhibits reduced spatial sparsity in the error surrounding the model.
This interprets as a more uniform scan of the target when seeking to update visual vocabularies that are defined for a smaller view range. This means that for a fixed number of words in each vocabulary, each view has less opponents to be compared with and the resulting visual words are more local to that region. While each view has a higher chance to represent a certain view range in the vocabulary, there is a lower chance for its similar views to be accepted into the set, as the dominant local features have already been identified. This effect \textcolor{black}{was} reinforced when the number of visual words per each vocabulary in the BoV \textcolor{black}{was} increased. The higher the number of visual words attributed to each view range, the higher the chance of familiarity of the newly captured view and details exposed to it.
\vspace{-1em}
\subsection{Appearance-based Next-Best-View Planning: Baseline Comparison and Generalization}
Based on the problem formulation in Sec.~\ref{rl} and the training algorithm in Sec~\ref{training}, we trained an agent in a reinforcement learning cycle depicted in Fig.~\ref{nbvframework} to scan multiple structures from the dataset we generated using VPT and publicly available 3D models. 

\begin{figure}[t]
\centerline{\includegraphics[width=\columnwidth]{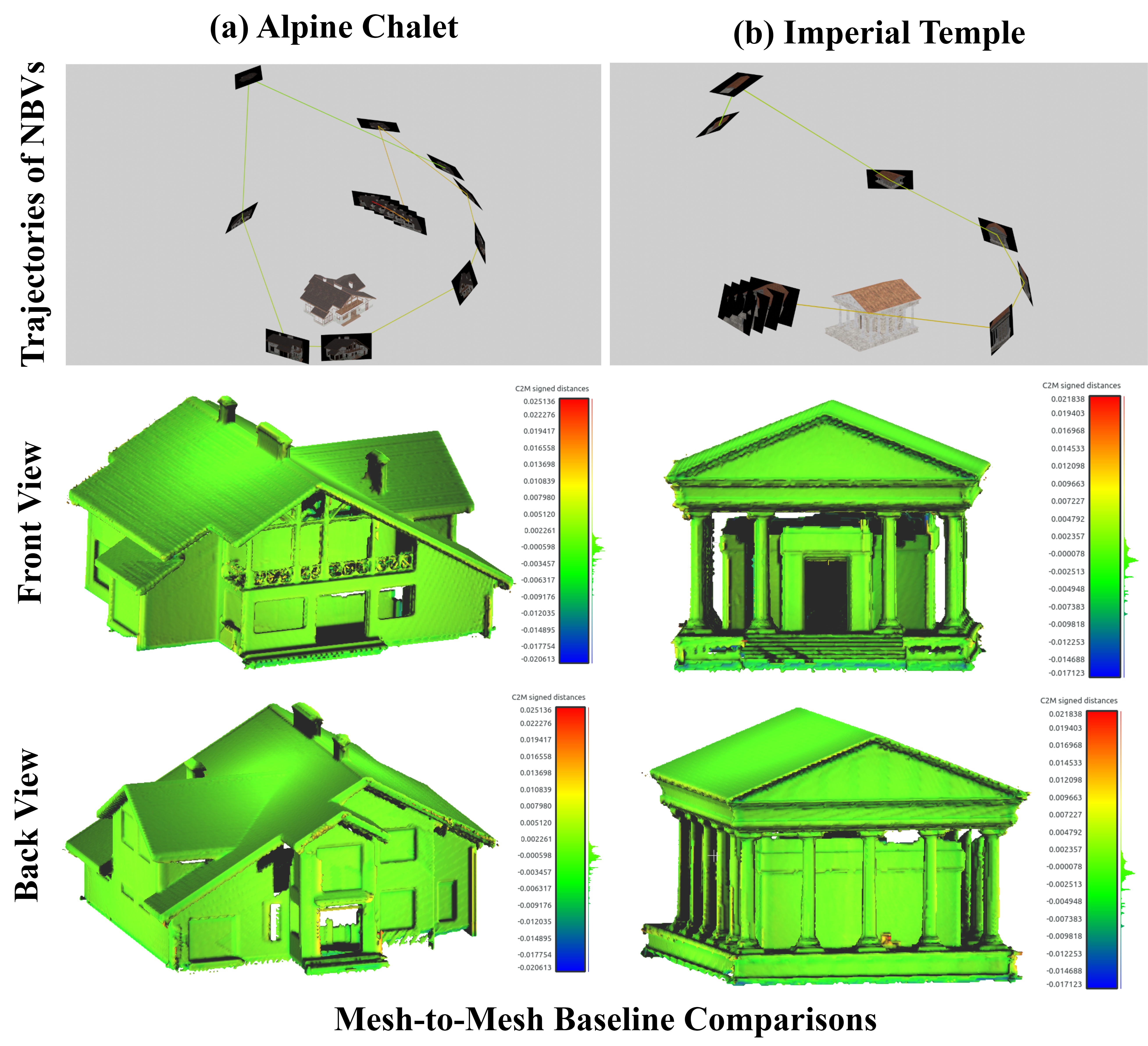}} 
\caption{Testing the generalizablity of our RL-based NBV planner on two samples from our dataset. The resulting Hausdorff distance and Chamfer discrepancy of the \textit{Alpine Chalet} reconstruction were calculated to be $9.47$ cm and $1.01$ cm while those of the \textit{Imperial Temple} were $6.35$ cm and $0.82$ cm, respectively.}
\label{testing}
\vspace{-1.4em}
\end{figure} 
We set the termination condition as the completion of one pass around the target to demonstrate the ability of the model to iteratively find the optimal views, and \textcolor{black}{compared} the resulting reconstructions with the results of a full coverage scan of the target as the baseline. The resultant trajectory for a sample from the dataset is shown in Fig.~\ref{evalfig}.\textcolor{black}{ Furthermore, we applied the policy trained on a single structure in the initial phase to evaluate its performance on two unseen buildings. This allowed us to assess the policy's ability to generalize across different structures and settings.} Shown in Fig.~\ref{testing}, the results demonstrate the efficacy of the model in finding the optimal views which result in high-quality reconstructions. \textcolor{black}{An average runtime of 0.893 seconds was measured while using an NVIDIA RTX 3060 with up to 16 GB of memory for the evaluation and testing of the model. This planning execution time is only a small fraction of the time required for scanning of an object in real-world conditions (i.e., on the order of several minutes).
The implementation is thus practical with either an onboard GPU or a ground-based GPU calculating and communicating the mission with a UAV.}
\vspace{-0.8em}
\section{Conclusions and Future Work}\label{conclusion}
Our study tackled the challenge of model-free view planning by introducing a novel appearance-based computational representation of reconstruction targets that can be of utmost utility for UAV-based aerial photogrammetry. Our model enables utility assignment to the views without tracking a full or partial reconstruction of the target through tracking unfamiliar visual features with its vocabularies contained within what we called a Bag-of-Views (BoV). We also developed the View Planning Toolbox (VPT), which offers a comprehensive solution for training, evaluation, and custom dataset generation in the context of view planning and 3D reconstruction. Our first set of experiments focused on exploring the size of BoV and the vocabularies within it on reconstruction quality. Using the reconstruction results from a complete coverage scan of the target as a baseline, we found that the BoV model achieved a remarkable reduction of views used for reconstruction ($70.6\%$ decrease) while simultaneously reducing the reconstruction error ($33.5\%$ decrease). These outcomes showcased the efficacy of our model in identifying optimal views for reconstruction. Building upon this proof of concept, we extended the application of the BoV model to shape the reward of a reinforcement learning (RL) agent trained using the Soft Actor-Critic (SAC) algorithm for online NBV planning. Once again, our model yielded high-quality reconstructions with a significantly low number of views (down to $5\%$ of the number of baseline views). Furthermore, the RL model exhibited substantial generalizability to unseen targets. Notably, we discovered that the degree of generalizability depended on the relative visual complexity of the training and testing environments, further validating the effectiveness of our appearance-based view selection approach. 
\textcolor{black}{We emphasize that our choice of simulated data was deliberate, allowing us to present a comprehensive and novel approach unencumbered by real-world data complexities. Promising future research can include using custom feature extractors and pre-training the visual vocabularies of the BoV for tracking certain visual features associated with structural defects in infrastructure as well as exploring real-world data evaluation. While we used SIFT feature extractor for creating BoV models which can cause vulnerability to environmental and lighting conditions, custom feature extractors can be separately developed to introduce robustness to such conditions.}
 \bibliographystyle{elsarticle-num} 
 \bibliography{main}





\end{document}